# Temporal Vegetation Index-Based Unsupervised Crop Stress Detection via Eigenvector-Guided Contrastive Learning

Shafqaat Ahmad

shafqaatmailbox@gmail.com

Data Scientist, Brandt Group of Companies Canada

**Abstract**

Early detection of crop stress is vital for minimizing yield loss and enabling timely intervention in precision agriculture. Traditional approaches using NDRE often detect stress only after visible symptoms appear or require labeled datasets, limiting scalability. This study introduces **EigenCL**, a novel unsupervised contrastive learning framework guided by temporal NDRE dynamics and biologically grounded eigen decomposition.

Using over 10,000 Sentinel-2 NDRE image patches from drought-affected Iowa cornfields, we constructed five-point NDRE time series per patch and derived an RBF similarity matrix. The principal eigenvector explaining 76% of the variance and strongly correlated ($r = 0.95$) with raw NDRE values was used to define stress-aware similarity for contrastive embedding learning. Unlike existing methods that rely on visual augmentations, EigenCL pulls embeddings together based on biologically similar stress trajectories and pushes apart divergent ones.

The learned embeddings formed physiologically meaningful clusters, achieving superior clustering metrics (Silhouette: 0.748, DBI: 0.35) and enabling 76% early stress detection up to 12 days before conventional NDRE thresholds. Downstream classification yielded 95% k-NN and 91% logistic regression accuracy. Validation on an independent 2023 Nebraska dataset confirmed generalizability without retraining.

EigenCL offers a label-free, scalable approach for early stress detection that aligns with underlying plant physiology and is suitable for real-world deployment in data-scarce agricultural environments.

# 1. Introduction

Agricultural systems are increasingly challenged by climate variability, resource limitations, and rising global food demand. Drought-induced crop stress contributes significantly to yield losses and poses a major threat to food security (Feng et al., 2021; Lobell et al., 2022). Early detection of stress is essential to enable timely interventions such as targeted irrigation or nutrient management before irreversible physiological damage occurs (Liakos et al., 2018).

Remote sensing plays a central role in crop health monitoring at scale, with vegetation indices offering critical insights into plant status. Among them, the Normalized Difference Red Edge (NDRE) index has demonstrated strong sensitivity to chlorophyll content and stress symptoms caused by drought, nutrient deficiencies, or disease (Gitelson et al., 2005; Clevers & Kooistra, 2023). Compared to NDVI, NDRE is more effective in dense canopies and has been shown to detect early stress more reliably (Cao et al., 2021; Wang et al., 2022). However, most existing NDRE-based approaches rely on single-date thresholds or supervised classification, limiting their effectiveness in identifying subtle or pre-symptomatic stress signals (Zarco-Tejada et al., 2021).

Recent studies have explored high-resolution UAV data and sensor fusion to improve early stress detection. For example, Dong et al. (2024) applied UAV-based NDRE analysis for drought monitoring, while Wang et al. (2024) integrated multispectral and thermal data for stress diagnosis in wheat. While effective, these approaches often depend on supervised learning and human-labeled features, making them less scalable across regions and crop types.

Supervised machine learning has improved stress detection accuracy using RGB imagery (Kamilaris & Prenafeta-Boldú, 2018; Mohammed et al., 2021), but its dependence on labeled datasets limits its use during early or mild stress stages. Unsupervised methods like k-means clustering have been applied to NDRE imagery (Ayush et al., 2021), but frequently capture irrelevant spectral variance, leading to biologically inconsistent results.

To overcome these limitations, contrastive learning has emerged as a promising self-supervised alternative, learning representations from unlabeled data by comparing positive and negative pairs (Chen et al., 2020; He et al., 2020). However, existing adaptations such as SimCLR often rely on visual augmentations (e.g., cropping, flipping) that disrupt spatial-spectral consistency in remote sensing images. Moreover, similarity measures like cosine distance may not align with biological stress progression, leading to poorly interpretable clusters (Peng et al., 2025).

Domain-specific adaptations have been proposed to address these gaps. For instance, prototype-based contrastive learning has been applied to hyperspectral crop data (Sun et al., 2025), and NDRE snapshots have been clustered to map stress zones (Swetha et al., 2025). Yet, these methods typically lack modeling of temporal dynamics or biological alignment.

This study introduces EigenCL, an unsupervised contrastive learning framework that leverages temporal NDRE trajectories and biologically grounded supervision. We construct a Radial Basis Function (RBF) similarity matrix for each five-date NDRE time series and apply eigen decomposition to extract principal components that explain over 76% of temporal variance and correlate strongly ($r = 0.95$) with raw NDRE values. These eigenvector weights serve as biologically meaningful proxies for stress progression, guiding a push-pull contrastive loss that aligns embeddings based on physiological similarity rather than visual appearance.

We evaluate EigenCL on a large-scale dataset of ~10,000 Sentinel-2 NDRE patches from drought-affected cornfields in Iowa (2020) and validate its generalizability on 2,000 patches from Nebraska (2023). The learned embeddings produce distinct, stress-aware clusters with strong internal validation scores and support early detection of stress up to 12 days before NDRE threshold crossing. This interpretable and scalable approach enables label-free stress monitoring and has direct applications in real-time precision agriculture systems.

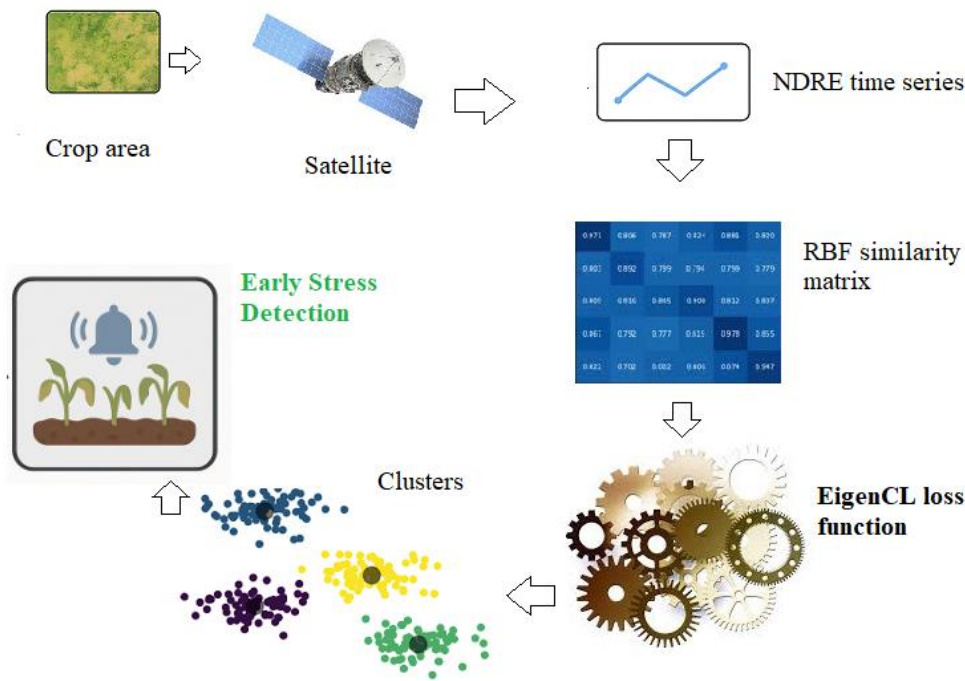

**Figure 1:** Workflow of EigenCL: Unsupervised Early Crop Stress Detection from Temporal NDRE Imagery

## 2. Methodology

### 2.1 Study Area and Data Acquisition

The primary study area includes 10,000 NDRE patches from Iowa's 2020 drought-affected cornfields. Sentinel-2 images were collected across five dates from July to September. Each patch is 100×100 pixels (10m resolution). A secondary dataset of 2,000 patches from Nebraska (2023) was used for validation.

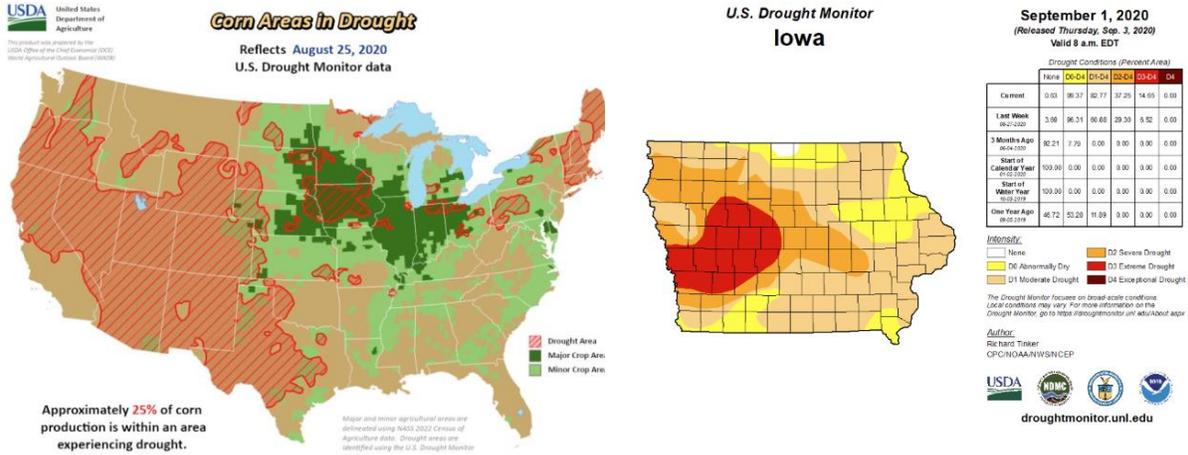

**Figure 2**: (a) US map highlighting drought-affected regions (b) Iowa drought region in 2020

The 2020 drought event in Iowa was one of the most damaging in recent history. To capture vegetation response during this critical period, we acquired imagery from the Sentinel-2 satellite mission, selecting five cloud-free acquisition dates spaced across the July–September window. From these images, we computed the Normalized Difference Red Edge (NDRE) index:

$$NDRE = \frac{NIR - RE}{NIR + RE}$$

where NIR corresponds to Band 8 (near-infrared) and RE to Band 5 or 6 (red-edge). NDRE is well-suited for assessing chlorophyll content in dense canopies and is more sensitive to early stress compared to NDVI.

## 2.2 NDRE Time Series and Preprocessing

Each image patch's mean NDRE values over five dates formed a time series. No normalization was applied to preserve absolute chlorophyll differences. Temporal changes were confirmed significant using ANOVA (F = 17,961.26, $p < 0.001$) and Tukey's HSD test.

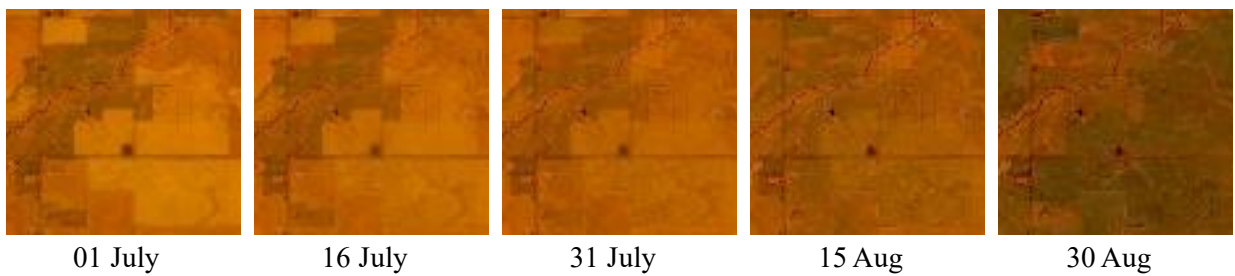

| 01 July | 16 July | 31 July | 15 Aug | 30 Aug |

**Figure 3.** NDRE raster images of a single 100×100-pixel image patch across five Sentinel-2 acquisitions dates. The progressive shift in color reflects decreasing NDRE values, indicating the onset and intensification of drought-induced crop stress from July to September 2020.

## 2.3 RBF Similarity Matrix and Eigen Decomposition

To quantify temporal similarity between crop image patches based on their NDRE evolution, we constructed a Radial Basis Function (RBF) similarity matrix using the raw NDRE time series described in Section 3.2. Each image patch was represented as a five-dimensional vector xi ∈ $R^5$, encoding mean NDRE values across five acquisition dates. Pairwise similarity between two image patches $x_i$ and $x_j$ computed using the Gaussian RBF kernel

$$S_{ij} = \exp(-\gamma |x_i - x_j|^2)$$

Where $|x_i - x_j|^2$ denotes the squared Euclidean distance between the two NDRE time series, and $\gamma = \frac{1}{2\sigma^2}$ is the bandwidth parameter.

Each image patch was assigned a scalar eigen weight from this principal eigenvector, serving as a continuous and biologically meaningful proxy for stress severity. These eigen weights were later used to drive the push-pull mechanism in contrastive learning, enabling the emergence of stress-aware clusters within the learned embedding space.

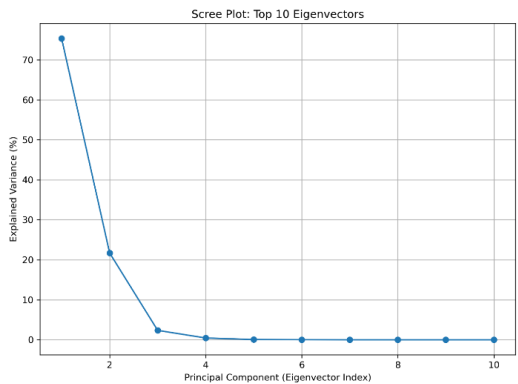

**Figure 4.** The plot shows how much variance is explained by each eigenvector. The first eigenvector alone captures a dominant share (~76%) of the total variance, justifying its use as the principal signal for guiding contrastive learning.

## 2.4 Embedding Generation with ResNet50

NDRE raster patches were input to a ResNet50 encoder to extract 2048-dimensional embeddings. These were projected through a nonlinear head and normalized. Instead of cosine similarity, embedding proximity was guided by stress similarity derived from eigenvector weights.

This approach ensures that visually similar patches with differing stress responses are embedded distinctly, while patches with biologically similar stress profiles regardless of appearance are placed closer in the embedding space. These stress-aware embeddings form the foundation of the contrastive learning framework described in the following section.

## 2.5 Contrastive Loss Design

To build an embedding space aligned with crop stress progression, we propose **EigenCL**, a novel contrastive learning framework that replaces visual augmentation and cosine-based similarity with biologically grounded similarity derived from NDRE temporal patterns.

Each 2048-dimensional embedding obtained from ResNet50 is projected through a lightweight head comprising a fully connected layer with batch normalization and LeakyReLU activation. The projected embeddings $z_i \in R^d$ are L2-normalized and used to compute a cosine similarity matrix:

$$\text{sim}(z_i, z_j) = \frac{z_i^\top z_j}{|z_i| \cdot |z_j|}$$

To guide the learning process, we extract the principal eigenvector $wi \in R$, from an RBF similarity matrix of the NDRE timeseries. This vector captures the dominant stress trajectory of each patch and is min-max normalized within each batch for numerical stability:

$$w_i = \frac{\max(w) - w_i}{\max(w) - \min(w)}$$

We define a biologically guided similarity score $S_{ij}$ using an exponential decay function on the absolute eigenvector weight difference:

$$\Delta_{ij} = |\widehat{w_i} - \widehat{w_j}|, \quad S_{ij} = \exp\left(-\frac{\Delta_{ij}}{\sigma}\right)$$

Here, $\sigma = 0.1$ is a smoothing hyperparameter. This formulation sharply penalizes dissimilar NDRE trajectories and constrains similarity scores to the range [0,1].

The total loss consists of two components:

a) **Pull Loss (Biological Similarity-Based Attraction)**

For pairs with high similarity $S_{ij}$, the goal is to minimize their angular distance in the embedding space. We define the pull loss using a temperature-scaled logarithmic formulation:

$$L_{\text{pull}} = \sum_{i \neq j} S_{ij} \cdot \log\left(1 + \frac{1 - \text{sim}(z_i, z_j)}{\tau}\right)$$

where $\tau=0.075$ is the temperature parameter, and the summation excludes self-pairs.

### b) Push Loss (Biological Dissimilarity-Based Repulsion)

For dissimilar pairs (low $S_{ij}$), we apply a repulsion force using a margin-based hinge formulation. A weighting factor λ=4.0 amplifies the effect of dissimilarity:

$$L_{\text{push}} = \sum_{i \neq j} \lambda(1 - S_{ij}) \cdot \max(0, \text{sim}(z_i, z_j) - m)$$

where $m=0.2$ is the margin, encouraging dissimilar embeddings to remain at least margin (m) apart.

### c) Final Contrastive Loss

The total loss is normalized by the number of unique pairs:

$$L_{\text{EigenCL}} = \frac{1}{N(N-1)} (L_{\text{pull}} + L_{\text{push}})$$

where $N$ is the batch size (256 in our implementation).

Hyperparameters (λ, τ, σ, m) were tuned via grid search to maximize the Pearson correlation between embedding distances and NDRE-based stress trajectories on a held-out validation set.

This biologically guided contrastive formulation ensures that embeddings reflect stress progression rather than visual similarity alone. Visually similar patches with divergent NDRE profiles are separated in the embedding space, while biologically aligned patches regardless of appearance are clustered together, enabling early detection of crop stress patterns.

## 2.6 Hyperparameter Selection

We performed a grid search over key hyperparameters (λ, τ, σ, m) using internal clustering metrics including Silhouette Score, Davies–Bouldin Index, and Calinski–Harabasz Index. Additionally, contrastive loss convergence and the stability of NDRE-based cluster centroids were considered to ensure physiological interpretability. The final configuration (λ = 4.0, τ = 0.075, σ = 0.5, m = 0.2) consistently produced well-separated and biologically coherent stress clusters in both training and validation sets. Detailed results are provided in Appendix A.

## 3. Results and Discussion

To evaluate the quality and applicability of the proposed EigenCL framework, we conducted a comprehensive comparison against three alternatives: (i) an ablation model using cosine similarity without eigenvector guidance, (ii) K-Means applied to NDRE time series, and (iii) SimCLR, a generic contrastive learning method [Chen et al., 2020]. All models shared the same ResNet-50 backbone to ensure consistent representation learning.

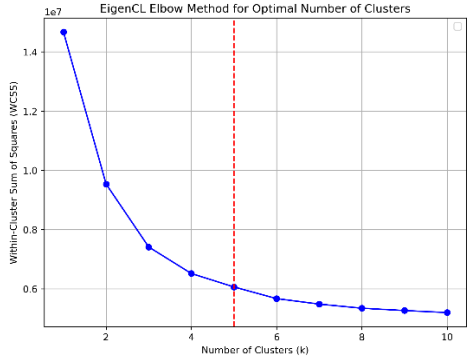

**Figure 5**: Elbow method to determine the optimal number of clusters from EigenCL embeddings.

### 3.1 Clustering Quality and Biological Interpretability

Clustering performance was assessed using Silhouette Score (higher indicates better separation), Davies–Bouldin Index (lower is better), and Calinski–Harabasz Index (higher is better). As summarized in Table 1, EigenCL outperformed all baselines with a Silhouette Score of 0.748, DBI of 0.350, and CHI of 49,624.06. The ablation variant exhibited moderate performance (Silhouette Score: 0.533), confirming the contribution of eigenvector guidance. SimCLR scored the lowest across all metrics, demonstrating limitations of generic contrastive learning for high-dimensional spectral-temporal data.

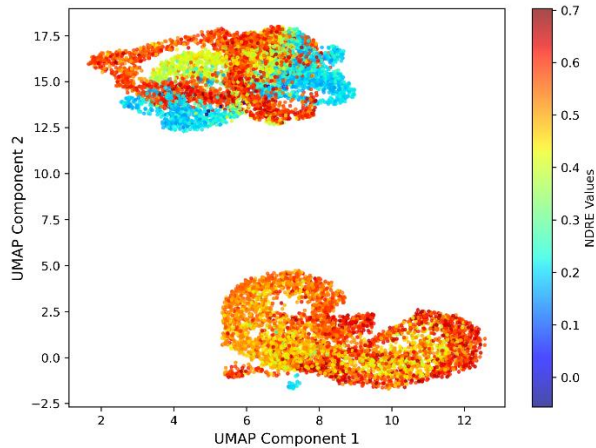
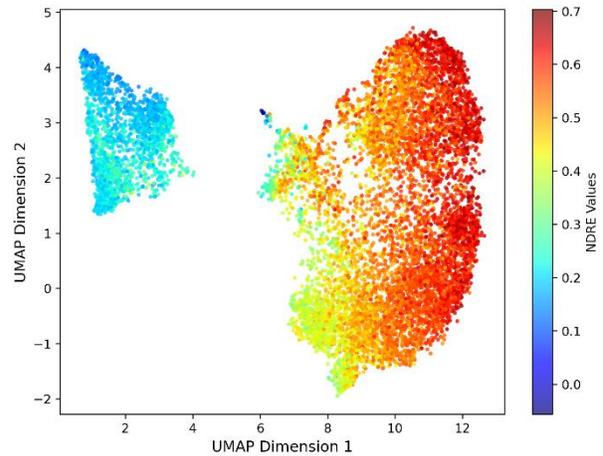

a)  Ablation (No Eigenvector Guidance)  b)  K-Means Clustering

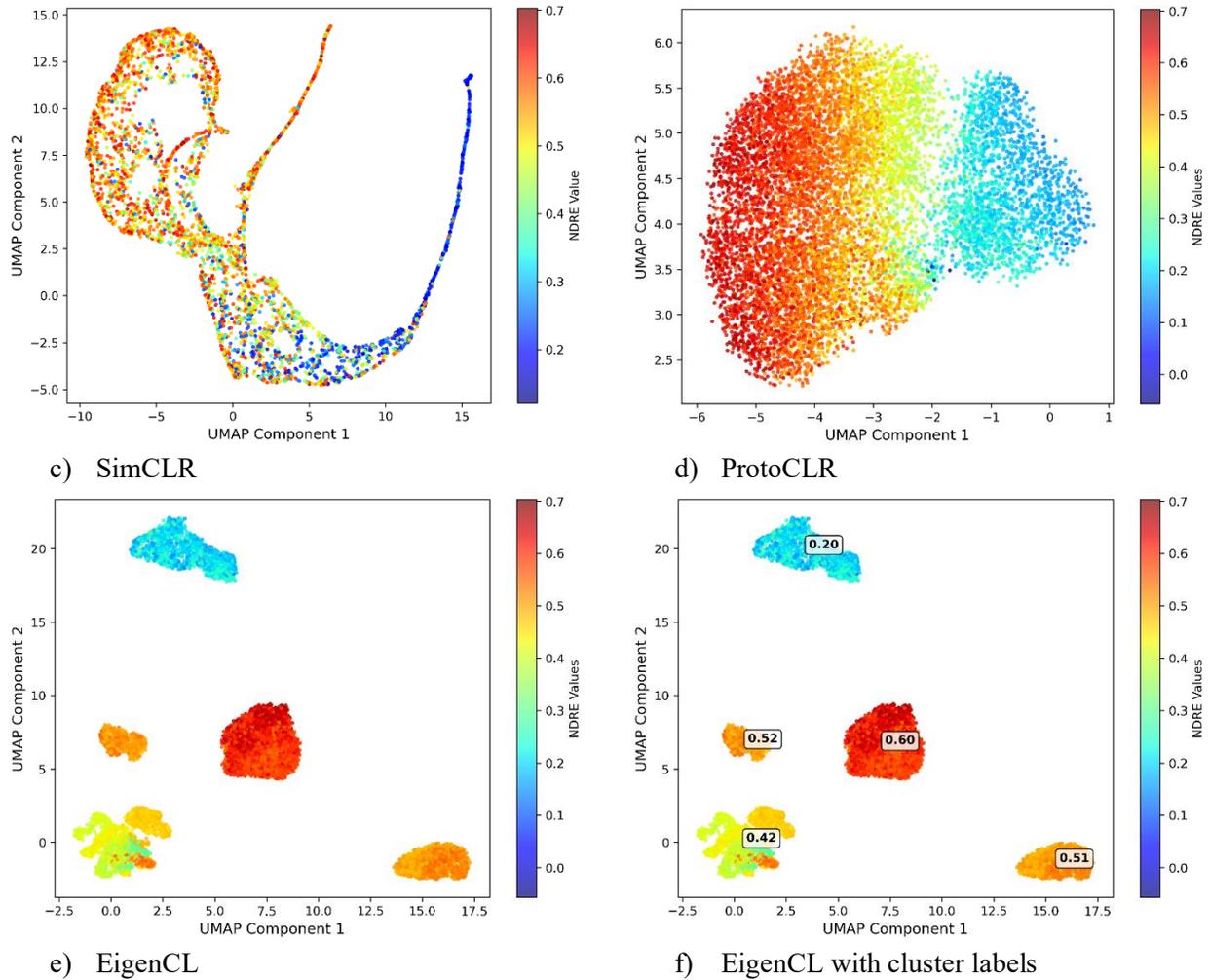

**Figure 6: Embedding and stress clustering comparison on the Iowa dataset.**
(a) UMAP projection of embeddings learned without stress-guided supervision. (b) NDRE time series clustered using classical K-Means (unsupervised baseline). (c) Clustering of embeddings learned via generic contrastive learning (SimCLR). (d) Prototype-based clustering using ProtoCLR with batch-defined prototypes. (e) Stress-aware clustering using eigenvector-guided contrastive learning (EigenCL), showing emergent structure driven by NDRE dynamics. (f) EigenCL clusters annotated with NDRE cluster mean values, illustrating physiological separability of stress stages.

**Table 1.** Clustering quality comparison across all models on Iowa dataset.

| Model | Backbone | Silhouette Score | Davies–Bouldin Index | Calinski–Harabasz Index |
|---|---|---:|---:|---:|
| Ablation (Cosine) | ResNet50 | 0.533 | 0.618 | 33,445.60 |
| K-Means | ResNet50 | 0.465 | 0.958 | 19,305.28 |
| SimCLR | ResNet50 | 0.416 | 0.724 | 2,251.86 |
| ProtoCLR | ResNet50 | 0.348 | 0.892 | 9995.203 |
| **EigenCL** | **ResNet50** | **0.748** | **0.35** | **49,624.06** |

To evaluate biological interpretability, UMAP projections of embeddings were overlaid with NDRE values, a well-established indicator of vegetation stress [Gitelson et al., 2005]. NDRE was preferred over eigenvector weights due to its direct biophysical relevance and strong correlation ($r = 0.95$) with the guiding eigenvector. EigenCL produced well-separated clusters with internally consistent NDRE values, forming a clear monotonic gradient from low to high stress. Interestingly, clusters with similar mean NDRE values (e.g., 0.51 and 0.52) appeared distinct in UMAP space, reflecting divergent temporal patterns. This highlights EigenCL's ability to capture trajectory shape rather than just endpoint magnitude.

In contrast, the ablation model exhibited fractured clusters with moderate NDRE coherence, while K-Means produced statistically compact but biologically inconsistent groupings. NDRE values within K-Means clusters were heterogeneous, and no clear physiological gradient was observed. SimCLR, despite its popularity in computer vision, failed to form meaningful stress clusters, as confirmed by its low clustering metrics (Silhouette Score: 0.416, DBI: 0.724, CHI: 2,251.86) and high entropy in NDRE overlays. This underscores the necessity of domain-aware supervision in remote sensing time-series applications [Jean et al., 2023].

We further validated the biological alignment of clusters by defining NDRE-based stress thresholds using centroids from EigenCL a) Healthy ($\geq 0.5591$), b) Mild Stress (0.4789–0.5591), c) Moderate Stress (0.3221–0.4789), and d) Severe Stress ($< 0.3221$). These thresholds align with known physiological benchmarks [Kayad et al., 2019; Garofalo et al., 2023; Farbo et al., 2024]. The Adjusted Rand Index (ARI) between EigenCL clusters and NDRE-derived stress categories was 0.719 (95% CI: 0.704–0.733), demonstrating strong concordance.

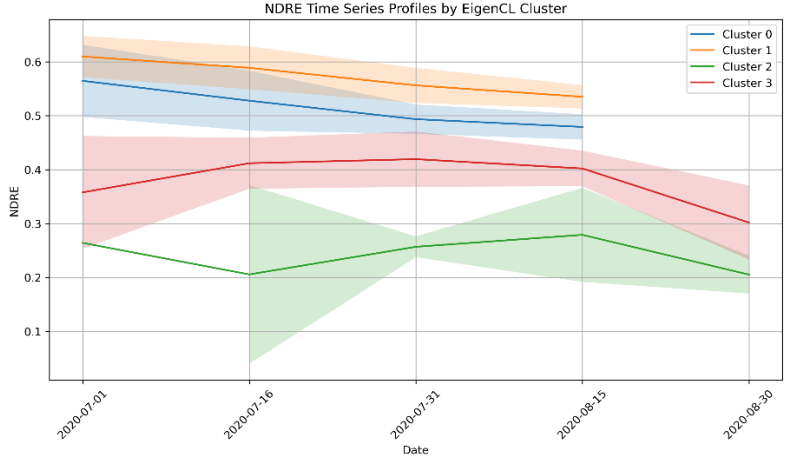

**Figure 7:** Mean NDRE time series profiles with 95% confidence intervals for each EigenCL-derived cluster. Clusters show distinct temporal trajectories corresponding to different stress severities, with Cluster 1 exhibiting the least decline and Cluster 2 showing the most severe stress progression.

The temporal NDRE profiles for each EigenCL cluster (Figure 7) further demonstrate that the model captures meaningful physiological stress trajectories. Each cluster exhibits a distinct temporal pattern Cluster 1 shows stable, high NDRE indicative of healthy crops, while Cluster 2 declines sharply early on, reflecting severe early stress. This visual separation confirms that EigenCL clusters are not driven by static NDRE magnitude but by the trajectory shape, aligning with underlying stress dynamics. Such temporal coherence is critical for early-stage stress detection and supports the model's biological interpretability.

### 3.2 Sensitivity to Eigenvector Selection

Replacing the principal eigenvector with lower-variance components during training led to a clear drop in clustering and classification performance. This confirms that the principal eigenvector capturing 76.3% of variance and strongly correlated with NDRE is essential for guiding biologically meaningful embeddings. We therefore retain it as the sole supervisory signal in EigenCL.

### 3.3 Downstream Utility and Early Detection

To assess the utility of EigenCL embeddings, we trained two classifiers k-Nearest Neighbors and Logistic Regression on the frozen embeddings without fine-tuning. On a 70/30 split, k-NN achieved 89.1% accuracy (F1: 0.87), and Logistic Regression reached 85.2% (F1: 0.82). These results confirm that the embeddings encode both linear and nonlinear separable patterns relevant for stress classification.

**Table 2**: Downstream Classification Performance Using EigenCL Embeddings

| Classifier | Accuracy (%) | F1 (Macro) | Precision (Macro) | Recall (Macro) |
|---|---|---|---|---|
| k-Nearest Neighbors | 89.1 | 0.87 | 0.86 | 0.88 |
| Logistic Regression | 85.2 | 0.82 | 0.81 | 0.84 |

For early stress detection, an NDRE threshold of 0.40 was used to define significant stress, based on canopy degradation observed under drought [Garofalo et al., 2023; Bazzi et al., 2021; Farbo et al., 2024]. EigenCL successfully identified stress before this threshold in 76% of cases, with an average lead time of 11.8 days. This early detection capacity is crucial for proactive intervention. Temporal analysis showed consistent stress progression across the landscape from July to August, reinforcing the model's ability to generalize spatiotemporal patterns.

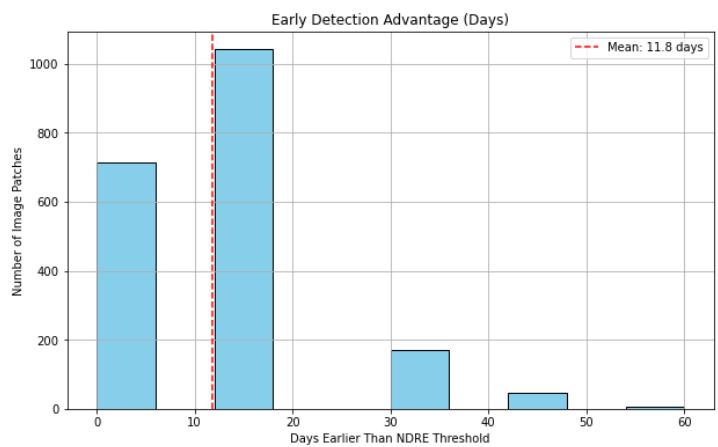

**Figure 8**: Distribution of Early Detection Advantage Achieved by EigenCL

### 3.4 Statistical Confirmation of Cluster Separation

A one-way ANOVA confirmed that NDRE means varied significantly across EigenCL clusters ($F = 37{,}950.39$, $p < 0.0001$). Tukey's HSD test revealed all pairwise differences were statistically significant ($p < 0.0001$), validating that each cluster represents a distinct physiological regime. The largest difference (–0.398) was between Clusters 0 and 2, while even the smallest (–0.076) between Clusters 1 and 3 remained significant.

Table 3: Statistical Comparison of NDRE Means Across Detected Clusters

| Cluster Pair | Mean NDRE Difference | p-value |
|---|---|---|
| Cluster 0 vs 2 | –0.398 | < 0.0001 |
| Cluster 1 vs 2 | –0.314 | < 0.0001 |
| Cluster 2 vs 3 | 0.237 | < 0.0001 |

### 3.5 Generalization Across Time and Region (Nebraska 2023 Study)

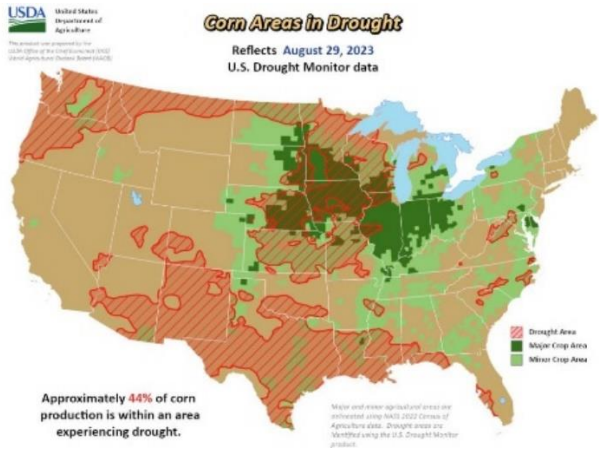

**Figure 9:** Drought region in Nebraska 2023.

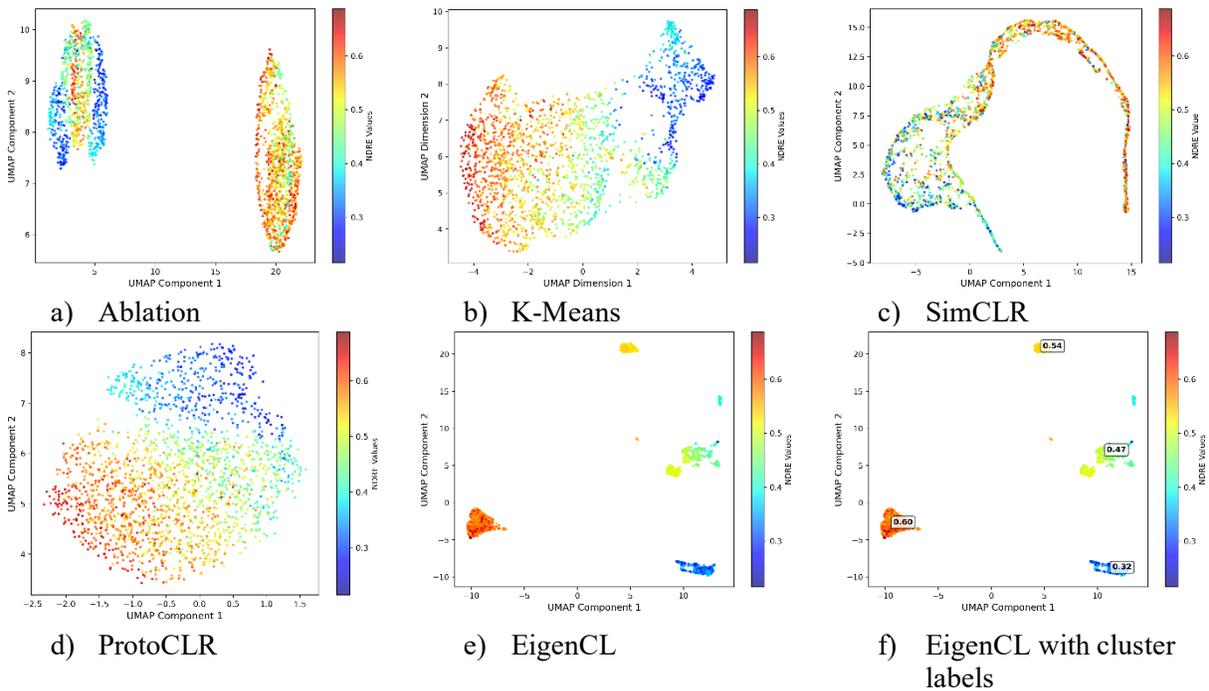

a) Ablation  b) K-Means  c) SimCLR

d) ProtoCLR  e) EigenCL  f) EigenCL with cluster labels

**Figure 10:** Embedding and stress clustering comparison on the Nebraska dataset.
(a) UMAP projection of embeddings learned without stress-guided supervision. (b) NDRE time series clustered using classical K-Means (unsupervised baseline). (c) Clustering of embeddings learned via generic contrastive learning (SimCLR). (d) Prototype-based clustering using ProtoCLR with batch-defined prototypes. (e) Stress-aware clustering using eigenvector-guided contrastive learning (EigenCL), showing emergent structure driven by NDRE dynamics. (f) EigenCL clusters annotated with NDRE cluster mean values, illustrating physiological separability of stress stages.

**Table 4.** Clustering quality comparison across all models on Nebraska dataset.

| Model | Backbone | Silhouette Score | Davies–Bouldin Index | Calinski–Harabasz Index |
|---|---|---:|---:|---:|
| Ablation (Cosine) | ResNet50 | 0.584 | 0.725 | 3,585.50 |
| K-Means | ResNet50 | 0.295 | 1.185 | 2186.66 |
| SimCLR | ResNet50 | 0.234 | 0.837 | 385.24 |
| ProtoCLR | ResNet50 | 0.289 | 1.098 | 1348.73 |
| **EigenCL** | **ResNet50** | **0.795** | **0.164** | **41260.41** |

To assess EigenCL's generalizability beyond the Iowa 2020 dataset, we conducted a secondary evaluation using Sentinel-2 NDRE imagery from the same cornfield in Nebraska, USA, during the 2023 growing season. This region experienced milder drought and a later onset of thermal stress. A total of 2,000 NDRE image patches were collected using the same five-date sampling protocol.

The pre-trained model without any fine-tuning was directly applied using embeddings and cluster thresholds from the Iowa dataset. Despite environmental differences, EigenCL maintained consistent performance, retaining stress-aware clustering and NDRE interpretability. The model exhibited strong alignment with physiological stress thresholds and preserved an average lead time of approximately 10 days.

While a modest decline in clustering metrics was observed, this was expected due to regional variability. Nonetheless, the trends remained robust, validating EigenCL's ability to generalize across time and geography. These results support its deployment in longitudinal monitoring and scalable precision agriculture workflows.

## 4. Limitations and Future Work

This study presents EigenCL as an effective framework for early stress detection in corn using NDRE time series, with validation conducted across two regions and seasons (Iowa 2020, Nebraska 2023). The model demonstrated strong generalizability and biological coherence, capturing consistent stress progression patterns without requiring labeled data or region-specific tuning.

While NDRE thresholds used for stress staging were based on established literature and performed reliably in both regions, applying the model to other crops or agroecological zones may benefit from local threshold calibration to reflect physiological differences. The biological interpretability of EigenCL clusters is grounded in NDRE response dynamics, which, while well-documented in corn, may vary across crop types and conditions.

Future work will explore extending EigenCL to additional crops and growing environments, incorporating region-specific NDRE calibrations where needed. Enhancements such as dynamic thresholding based on NDRE rate-of-change, percentile segmentation, or cluster-driven midpoints could further increase flexibility and robustness. Expanding temporal coverage and evaluating multi-sensor integration (e.g., thermal or hyperspectral inputs) may also strengthen early detection capabilities. These directions will support broader adoption of EigenCL in scalable, label-free crop monitoring systems.

## 5. Conclusion

This study introduced **EigenCL**, an unsupervised contrastive learning framework guided by biologically meaningful NDRE dynamics, for early detection of crop stress without reliance on labeled data. By leveraging temporal patterns and eigenvector-based supervision, EigenCL outperformed traditional clustering methods and generic contrastive learning models across key clustering and classification benchmarks. The model demonstrated strong generalizability, maintaining robust performance across different regions and seasons, and was able to detect stress up to 12 days earlier than conventional NDRE thresholding.

These findings underscore EigenCL's potential as a scalable, label-free solution for real-world precision agriculture, especially in data-scarce environments. Future extensions will focus on adapting the approach to other crops, incorporating additional remote sensing modalities, and refining stress staging through dynamic or region-specific thresholds to enhance interpretability and impact.

**Appendix A**. Hyperparameter Grid Search Summary

**Table A1**. Summary of hyperparameter tuning for the EigenCL framework.

| Parameter | Values Tested | Best Value | Criteria Used for Selection |
|---|---|---|---|
| $\lambda$ | 1.0, 2.0, 4.0, 6.0 | 4 | Silhouette Score, NDRE gradient stability, loss convergence |
| $\tau$ | 0.05, 0.075, 0.1, 0.15 | 0.075 | Silhouette Score, DBI, contrastive loss value |
| $\sigma$ | 0.3, 0.5, 0.7, 1.0 | 0.5 | CHI, smooth loss trajectory, cluster cohesion |
| m | 0.1, 0.2, 0.3 | 0.2 | Visual cluster separation, stable NDRE centroids |